# Artificial Intelligence and Asymmetric Information Theory

Tshilidzi Marwala and Evan Hurwitz

tmarwala@gmail.com, hurwitze@gmail.com

University of Johannesburg

**Abstract**

When human agents come together to make decisions, it is often the case that one human agent has more information than the other. This phenomenon is called information asymmetry and this distorts the market. Often if one human agent intends to manipulate a decision in its favor the human agent can signal wrong or right information. Alternatively, one human agent can screen for information to reduce the impact of asymmetric information on decisions. With the advent of artificial intelligence, signaling and screening have been made easier. This paper studies the impact of artificial intelligence on the theory of asymmetric information. It is surmised that artificial intelligent agents reduce the degree of information asymmetry and thus the market where these agents are deployed become more efficient. It is also postulated that the more artificial intelligent agents there are deployed in the market the less is the volume of trades in the market. This is because for many trades to happen the asymmetry of information on goods and services to be traded should exist, creating a sense of arbitrage.

**Introduction**

The theory of asymmetry of information and its impact in the markets has been extensive (Aboody and Lev, 2000). Some of the undesirable consequences of information asymmetry are moral hazards, monopoly of information and adverse selection (Aboody and Lev, 2000). The difference in perception of value of goods and services is the basis of trade. Humans for much of history have been agents that execute trade. Humans often have a distorted perception of information for all kinds of reasons and often change their minds about one thing or another. There is a thought experiment where a human being is given 20 paintings to rank in the order of how valuable they are. Ten days later the same human beings are given the same paintings and often it is found that the rankings differ markedly from the first instance. The same phenomenon governs human beings when they participate in the market. Consequently, the character of the market is deformed simply because of the inconsistency of the human agents who are participating in the market. This is primarily because people have a warped sense of information and therefore their decisions are bounded rationally.

Over the last few years much of the decisions in the market are being made in increasing numbers by intelligent machines. These intelligent machines are able to analyze vast amounts of information, are able to identify latent information which are not easily accessible to humans through the advent of big data and deep learning and are in effect flexibly rationally bounded to better decision making due to advances in signal processing techniques, missing data estimation methods and the ability to process vast amount of data fast (Marwala, 2015; Simon, 1957).

**Introduction to Asymmetric Information**

Information asymmetry is a study of decisions made by human beings were one human agent has more information than another human agent. There are cases were information asymmetry is not desirable, for example in an interview settings were one human agent (the potential employer) needs to know as much as possible about the potential employee and this problem was studied extensively by Nobel Laureate Michael Spence (Spence, 1973). In this case the potential employer who is interested in knowing as much information about the employee signals to the employee to reveal as much information as possible. Likewise the potential employee will signal to the potential employer information such as qualifications thus sending a message that he/she is skilled in the job. How does artificial intelligence help resolve such asymmetry of information through signaling? How does big data and deep learning resolve this asymmetrical dilemma? Of course these days, social networks which are often enabled with artificial intelligence are able to signal information much more accurately than a human agent is able to do. Therefore, artificial intelligence is able to help resolve asymmetry of information.

The other issue with information asymmetry is that of screening which was studied by Nobel Laureate Joseph Stiglitz (1974). Stiglitz introduced the concept of screening where the human agent that knows less information induces the human agent that knows more information to reveal some information. With the advent of artificial intelligence it is no longer as necessary for one human agent to try to induce another human agent to reveal more about itself. One human agent can be able to use the internet to create a profile of the other human agent which is more accurate and informative than the information that could have been extracted from the party in question. This is due to the fact that human agent forgets facts very easily and may even not be capable of revealing all the information for all sorts of reasons.

**Artificial Intelligence**

Artificial Intelligence is a computational approach which is motivated by natural intelligence such as the functioning of the brain, the organization of animals as well as the evolution of physical and biological systems (Russell and Norvig, 2010). Artificial intelligence is therefore capable of learning, adapting, optimizing and evolving. Some examples of artificial intelligence techniques are neural networks which are capable of learning from data, genetic algorithm which is capable of evolving and ant colony optimization which is capable of optimizing. Artificial intelligence has been successfully used in decision making in a number of areas such as in engineering (Marwala, 2010&2012; Marwala et al, 2016), in missing data estimation (Marwala, 2009), in economics (Marwala, 2013), in political science (Marwala, 2011) as well as in rational decision making (Marwala, 2014&2015). More recently artificial intelligence has been applied in the markets for high frequency trading.

**Efficient Market Hypothesis**

Efficient market hypothesis is theory proposed by Nobel Laureate Eugene Fama which states that the market incorporates all the information such that it is impossible to beat the market (Fama, 1965). It thus follows that the only way to beat the market is to engage in high risk transactions. Implicit in the efficient market hypothesis is the fact that the agents that participate in the market are rational. Of course we now know that human agents are not rational and therefore the markets cannot be rational. Theories such as prospect theory and bounded rationality have proven that at best human agents are not fully rational but almost always are not rational (Simon, 1974; Kahneman and Tversky, 1979). Marwala (2015) surmised that artificial intelligent agents make markets more rational than human agents.

**Asymmetric Information and Trading Volume**

Suppose we have a market where all the transactions are conducted by machines that are capacitated by artificial intelligence techniques. In this scenario, the artificial intelligence machine will look at all the information at its disposal including information available in the internet to make decisions. In this scenario, the degree of rationality in the market is increased because an irrational agent i.e. a human being is not participating in the market. In this situation the degree of information asymmetry in the markets will be greatly reduced almost to no asymmetry. Asymmetry in many trading scenarios is in fact a driver in the trading process, creating (on both sides of the trade) a sense that each party is getting a better deal than the other. Consider a case in which a given commodity is being traded, and party A (the seller) considers the good worth $10, while party B (the buyer) considers the good worth $15. Through bargaining, they agree to a sale price of $13, with party A believing himself to have made a profit of $3, and party B believing himself to have made a profit of $2. This situation cannot arise if both parties have access to perfect information as to the true value of the underlying commodity, and the trade then does not take place. In this situation the number of transactions that will happen in this market will be greatly reduced because there is no information asymmetry to be exploited to make money.

**Asymmetric Information and Market Efficiency**

Another aspect that warrants close study is the relationship between asymmetric information and market efficiency. If a human agent A has smaller amount of information to make a decision than another human agent B, then the decision of human agent A is more rationally bounded than the decision of human agent B. If a market is full of agents with the same characteristics as those of agents A and B then such a market cannot be efficient because the decisions of significant players of the market are based on limited information. Therefore, asymmetric information even though they promote trading make markets inefficient because they distort the markets. If human agents A and B are replaced by autonomous artificial intelligent agents A and B, then the information that each agent can be able to mine in the cyberspace will be similar especially if their capabilities are assumed to be the same. This then means the information that is at the disposal of artificially intelligent agents A and B are symmetrical. If a market is full of agents such as the artificially intelligent agents A and B then the market will have agents where information is more symmetrical and therefore it will be more rational. Moreover, these artificially intelligent agents will be able to analyze all the data at their disposal, estimate latent information and process all the information at their disposal than a human being. Thus the decisions of the artificially intelligent agents will be less rationally bounded than the decisions of the human agents. Therefore, the deployment of artificial intelligent agents make information in the markets more symmetrical (or less asymmetrical) and this in turn makes the markets more efficient.

**Conclusions**

This paper has proposed that the degree of asymmetry of information between two artificial intelligent agents is less than that between two human agents. As a result, it is also postulated that the more artificial intelligent there is in the market the less is the volume of trades in the market, and the overall efficiency of the market is likely to improve over time as the market becomes more saturated with intelligent trading and analysis agents.